%% file: main.tex
\pdfoutput=1
\documentclass[runningheads]{llncs}
\PassOptionsToPackage{table}{xcolor}

\usepackage{eccv}

\usepackage{eccvabbrv}
\usepackage{graphicx}
\usepackage{booktabs}
\usepackage{longtable}
\usepackage{multirow}
\usepackage{array}
\newcolumntype{C}[1]{>{\centering\arraybackslash}p{#1}}
\newcolumntype{R}[1]{>{\raggedleft\arraybackslash}p{#1}}
\usepackage{amsmath,amssymb}
\usepackage[numbers,square,comma]{natbib}
\usepackage{pdflscape}

\usepackage[accsupp]{axessibility}

\usepackage[pagebackref,breaklinks,colorlinks,citecolor=eccvblue]{hyperref}

\usepackage{orcidlink}

\usepackage{xcolor, colortbl}
\definecolor{groupbg}{HTML}{E8EEF7}
\definecolor{dashgray}{HTML}{A8A8A8}
\newcommand{\best}[1]{\textbf{#1}}

\newcommand{\nr}{\textcolor{dashgray}{--}}
\definecolor{arrowclr}{HTML}{049BE5}  
\newcommand{\up}{\,\textcolor{arrowclr}{$\uparrow$}}
\newcommand{\dn}{\,\textcolor{arrowclr}{$\downarrow$}}

\begin{document}

\title{Post-Training in End-to-End Autonomous Driving: A Unified View}

\titlerunning{Post-training in End-to-End Autonomous Driving: A Unified View}

\author{Ruining Yang\inst{1}\thanks{Equal contribution. Email: \texttt{yang.ruini@northeastern.edu, wang.muxin@northeastern.edu}} 
\and Muxing Wang\inst{1}$^{\star}$
\and Yixiao Chen\inst{1}
\and Tongfei Guo\inst{1}
\and Yi Xu\inst{1}
\and \\
Can Cui\inst{2}
\and Zichong Yang\inst{2}
\and Yitian Zhang\inst{1}
\and Ziran Wang\inst{2}
\and Yun Fu\inst{1}
\and Lili Su\inst{1}
}

\authorrunning{R. Yang, M. Wang et al.}

\institute{Northeastern University, Boston, USA \and
Purdue University, West Lafayette, USA}

\maketitle

\begin{abstract}

End-to-end models that map multimodal inputs directly to future trajectories/maneuvers have emerged as an increasingly prominent research paradigm in autonomous driving. This class of models includes both Vision-Language-Action models and trajectory-generative planners.
Unlike classic machine learning applications, autonomous vehicles operate in safety-critical and interaction-intensive environments where traditional open-loop imitation of expert demonstrations is not sufficient to ensure reliability.
In particular, small execution errors can accumulate over time, while recovery behaviors are scarce in training data. 
In addition, long-horizon objectives such as safety and driving comfort are not captured by pointwise labels either.
These limitations have motivated a shift toward post-training techniques, which further refine driving policies beyond pure imitation.
This survey presents a unified view of post-training for autonomous driving by defining its scope and organizing the existing literature into four major families based on the form of supervision they use.
For each family, we discuss its capabilities, limitations, and open challenges. We aim to facilitate a systematic understanding of this emerging area and stimulate future research on reliable and efficient post-training for autonomous driving.
A collection of related papers is available at \url{https://github.com/RYNing/Awesome-Post-Training-In-Autonomous-Driving-Papers}.

\end{abstract}

\section{Introduction}
\label{sec:intro}

End-to-end models have become an increasingly prominent research paradigm in autonomous driving, wherein a single network takes multimodal inputs (e.g.\, camera images, navigation commands, and the vehicle's own state) and directly outputs future trajectories that instruct the maneuvers of the ego vehicle~\cite{hu2023planning,jiang2023vad,ma2025leapvad}.
Their growing popularity is driven by the promise of jointly optimizing perception, prediction, and planning,
reducing reliance on hand-engineered intermediate interfaces, learning richer cross-modal representations, and exhibiting stronger adaptability to complex driving scenarios.

Representative end-to-end driving models have evolved from early behavioral-cloning approaches that map raw pixels from front-facing camera and high-level navigation commands (e.g.~turn left/right) directly to low-level controls~\cite{bojarski2016end,codevilla2018end}, to trajectory-generative planners that predict future trajectories \cite{chitta2022transfuser,jia2023think,hdp2026,plannerrft2026,driveanchor2026,eponav2_2026}, and Vision-Language-Action (VLA) models that leverage advanced
multimodal representations and language reasoning to improve scene understanding and decision making~\cite{alpamayo2025,autodriver2_2025,autodrivep3_2026,reasoningvla2025}. Despite the architectural differences among these approaches, they largely rely on imitation learning (i.e., supervised learning on offline training data) to produce coarse actions or trajectories that mimic expert driving behavior from demonstrations.

\begin{figure}[t]
    \centering
    \includegraphics[width=\linewidth]{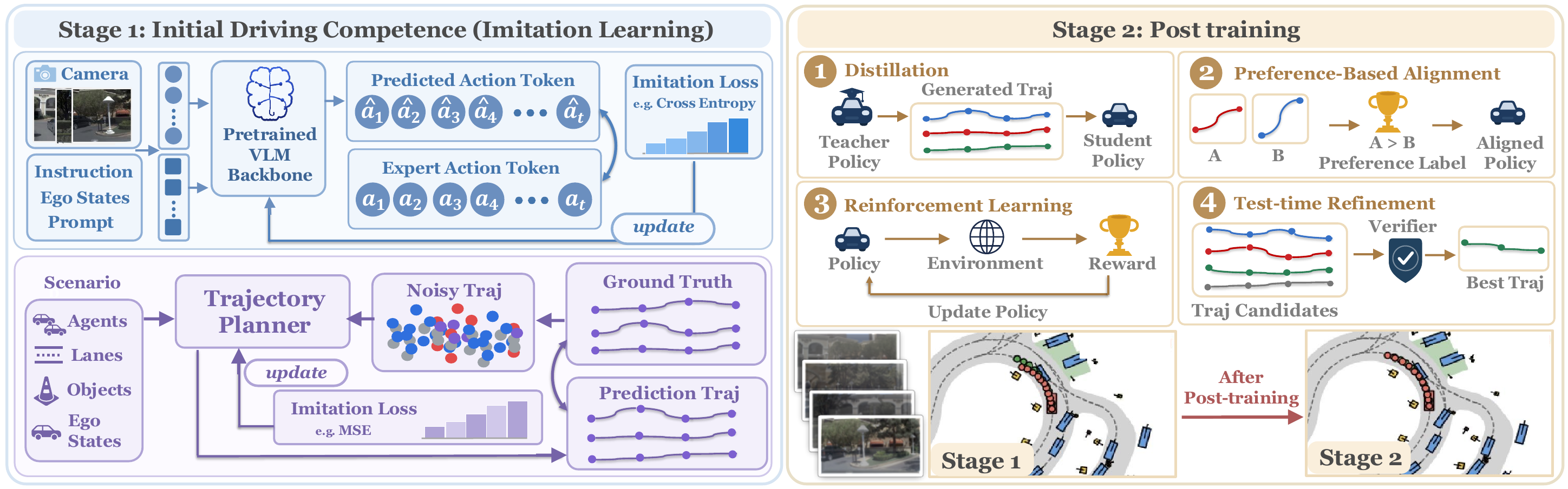}
    \caption{
    Two-stage training paradigm for autonomous driving policies. Stage 1 builds initial driving competence through imitation learning, while Stage 2 further improves the policy through post-training.}
    \label{fig:fig1}
    \vspace{-0.4cm}
\end{figure}

Unlike classic machine learning applications~\cite{yao2026bipartite,yang2026tp}, autonomous vehicles operate in safety-critical and interaction-intensive environments where traditional open-loop imitation is not sufficient to ensure reliable driving behaviors. 
In closed-loop executions, where a model's actions influence future observations, even small errors can compound over time; for example, as shown in Fig.\,\ref{fig:fig1}, a minor lane-centering deviation may gradually accumulate and eventually drive the vehicle outside the drivable area.  
Moreover, many important aspects of driving quality are not well captured by pointwise imitation losses: two trajectories may exhibit similar displacement error with respect to an expert demonstration, yet one may involve unnecessary hard braking and acceleration while the other is smooth. These limitations highlight the need for supervision beyond offline demonstrations, which we refer to as {\em post-training} in this survey.

Recent progress in driving-specific feedback and evaluation has made post-training increasingly viable for autonomous driving. Rule-based driving scores~\cite{navsim2024}, Vision-Language-Model (VLM) critics~\cite{drivecritic2025}, human-takeover logs~\cite{spanvla2026}, world models~\cite{explorevla2026,dreamerad2026}, and closed-loop~\cite{bench2drive2024,omnidreams2026} or pseudo-closed-loop simulators~\cite{navsimv2_2025} provide ways to assess safety, comfort, progress, interaction, and rule compliance beyond expert demonstrations.
These developments have led to a fast-growing but fragmented literature, which this survey aims to organize around the post-training refinement pipeline. Section~\ref{sec:pipeline} first defines post-training and clarifies how it differs from initial imitation learning and learning from scratch. Sections~\ref{sec:distill}--\ref{sec:testtime} then review four major families of post-training methods: distillation, preference-based alignment, reinforcement learning, and test-time refinement. Section~\ref{sec:evaluation} discusses evaluation protocols and challenges, and Section~\ref{sec:future} outlines open problems and future directions.

Most existing surveys cover end-to-end driving \cite{yang2025survey,xu2026survey}, VLA models \cite{hu2025vision,jiang2025survey}, foundation models \cite{gao2025survey}, and closed-loop training \cite{karkus2025beyond}, but post-training is less often treated as a distinct stage. Our contributions are three-fold:

\begin{itemize}
\item \textbf{A unified definition and taxonomy.} We define post-training as a distinct stage, distinguish it from initial training and learning from scratch, and organize the literature into four families by the form of supervision: distillation, preference-based alignment, RL, and test-time refinement.
\item \textbf{A structured review of all four families of supervision.} For each family, we examine how the supervision is constructed, how training data or rollouts are collected, and how the policy or its outputs are improved.
\item \textbf{A discussion of evaluation challenges and open problems.} We summarize the main difficulties in evaluating post-trained policies, including benchmark saturation, coarse closed-loop metrics, limited real-vehicle evaluation, and inference cost, and we outline future directions.
\end{itemize}

\section{Post-training: Definition and Families}
\label{sec:pipeline}
This section defines post-training and presents a unified taxonomy of its major methodological families. For expositional simplicity, we use {\em imitation learning} to refer broadly to the initial supervised training  on offline demonstrations, recognizing that practical training pipelines may incorporate additional components~\cite{chitta2022transfuser,hu2023planning,jiang2023vad}. We then define post-training as any subsequent refinement of a pretrained model using supervision beyond this initial training stage. 

\vspace{0.5em}
\noindent{\bf Imitation learning.}  
Given an offline dataset \(\mathcal{D}_{\text{IL}}=\{(o_i,y_i)\}_{i=1}^{N}\), where \(o_i\) denotes a driving observation and \(y_i\) denotes the corresponding expert target (e.g.~an action or a future trajectory), the initial policy \(\pi_{\theta_0}\) is trained in a supervised manner by solving 
\vspace{-0.3em}
\begin{align}
\label{eq: imitation learning}    
\min_{\theta}~ \frac{1}{N} \sum_{(o_i, y_i)\in \mathcal{D}_{\text{IL}}} \mathcal{L}_{\mathrm{IL}}(y_i,\pi_\theta(o_i)),
\end{align} 
\vskip -0.3\baselineskip 
\noindent 
where $\mathcal{L}_{\mathrm{IL}}$ is the loss that measures the discrepancy between the model prediction and the corresponding expert target. We refer to the resulting policy \(\pi_{\theta_0}\) as the imitation-learned policy.

For traditional model architectures (i.e., the one without foundation model backbones), \(y_i\) is usually represented as continuous waypoints or trajectories, and the policy is optimized with a trajectory regression loss~\cite{zheng2025diffusion,liao2025diffusiondrive}. 
For the end-to-end models that contain foundation model backbones, 
\(y_i\) is often represented as expert action or trajectory tokens, and the policy is optimized with token-level cross-entropy loss~\cite{mao2023gpt,hwang2024emma,autovla2025}. Furthermore, when foundation model backbones are employed, the objective in Eq.~\eqref{eq: imitation learning} is typically not optimized from scratch. Instead, the pretrained model is adapted to the driving domain through few-shot fine-tuning or other parameter-efficient adaptation techniques.

\vspace{0.5em}
\noindent{\bf Post-training.}  
Post-training is any subsequent refinement stage applied to this imitation-learned policy. 
It seeks to improve closed-loop behavior by mitigating error accumulation and enhancing safety, comfort, and other aspects of driving quality.  
However, these objectives cannot generally be reduced to a universal and fully specified optimization problem amenable to standard optimization tools. 
Instead, they are often conveyed through heterogeneous forms of supervision, such as privileged teacher policies, preference comparisons, reward signals, or test-time feedbacks. 

Despite their diversity, a large class of post-training methods can be viewed through a common abstraction:  
\vspace{-0.2em}
\begin{align}
\label{eq: 1-3 post-training}
   \min_{\theta \in B(\theta_0)}\; \mathbb{E}_{(\tilde{o},\,\text{signal}) \sim \mathcal{D}_{\text{PT}}}\Bigl[ \mathcal{L}_{\text{PT}}\bigl(\pi_\theta(\tilde{o}),\; \text{signal}\bigr) \Bigr] + \lambda\,\Omega(\theta),
\end{align}
\vskip -0.3\baselineskip 

\noindent where $B(\theta_0)$ denotes the pre-specified neighborhood around the imitation policy $\theta_0$ within which we refine the policy;  
$\tilde{o}$ is the post-training observations (which may differ from those in $\mathcal{D}_{\text{IL}}$) that are carefully designed or selected to reflect the desired post-training objectives (e.g.\,driving comfort and safety);  
and $\text{signal}$ denotes the supervision signal, whose specific form varies across different post-training methods. 
    Typically, \\
    -- \textit{(1) Distillation-based post-training (Section~\ref{sec:distill})} leverages a privileged teacher policy to provide dense behavioral guidance, such as recovery demonstrations to mitigate error compounding. Formally, the signal may take the form of a teacher policy $\pi_{\text{teacher}}(\tilde{o})$ or a teacher action $\hat{y}(\tilde{o})$;  \\
    --  \textit{(2) Preference-based post-training (Section~\ref{sec:preference})} uses relative comparisons between candidate behaviors to indicate which behavior is preferred. Formally, the signal is a preferred--rejected pair $(y^w, y^l)$;   \\     
    -- \textit{(3) Reinforcement learning (Section~\ref{sec:rl})} encodes richer guidance in scalar reward to sampled behaviors. Formally, the signal is $r(\tilde{o},y)\in \mathbb{R}$. 

    \vspace{0.3em} 
    
    \noindent The post-training loss $\mathcal{L}_{\text{PT}}$ likewise varies (e.g., KL divergence for distillation, DPO loss for preference, or a policy gradient/surrogate loss for RL) but these differences primarily reflect the optimization formulation rather than the source of supervision. Oftentimes, an addition regularization term $\lambda \Omega(\theta)$ is involved, where $\lambda$ is a regularization coefficient,
    and $\Omega(\theta)$ is a signal‑agnostic regularizer (e.g., distance to a reference policy such as $\text{KL}(\pi_\theta \| \pi_{\text{ref}})$).

\vspace{0.5em}
\noindent  -- \textit{(4) Test-time post-training (Section~\ref{sec:testtime}):} In contrast to Eq.~\eqref{eq: 1-3 post-training}, this family does not update the policy parameters. Instead, the fixed policy $\pi_{\theta_0}$ is refined through test-time selection or verification. Formally,  
\begin{align}
\label{eq:test_time_selection}
    y^* \;=\; \text{argmax}_{y \in \mathcal{Y}_{\text{cand}}(o)} \; v(o,y),
\end{align}
where $o$ is the driving observation, $\mathcal{Y}_{\text{cand}}(o)$ is a set of candidate outputs sampled from the fixed policy $\pi_{\theta_0}$, $v(o,y)$ is a verifier score that evaluates the quality of candidate $y$, and $y^*$ is the final output selected after test-time refinement.

\begin{figure}[t]
    \centering
    \includegraphics[width=\linewidth]{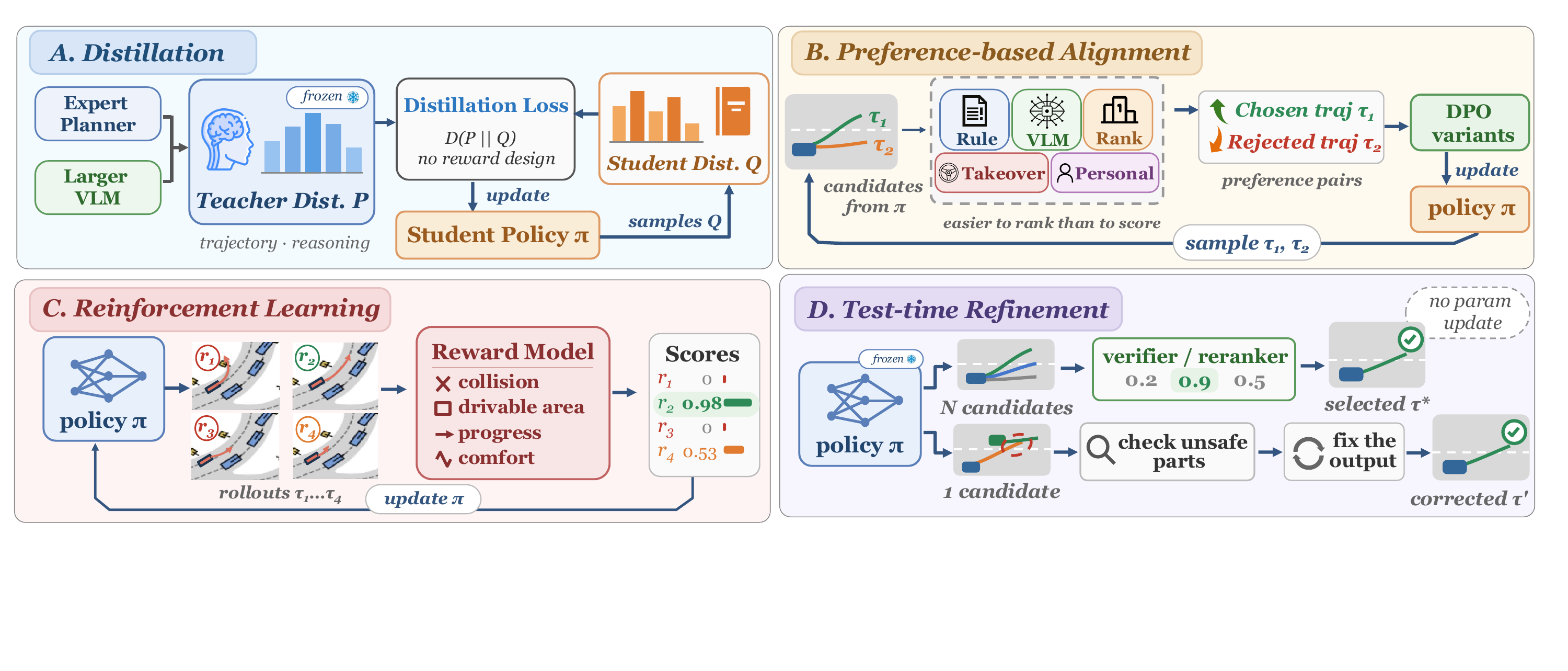}
    \caption{Overview of major post-training families for autonomous driving.}
    \label{fig:fig2}
    \vspace{-0.4cm}
\end{figure}

\section{Distillation}
\label{sec:distill}

Distillation refines $\pi_{\theta_0}$
using dense targets from a stronger teacher, such as an expert planner, a larger VLM, or a slowly updated copy of the policy itself.

Formally, distillation instantiates Eq.~(\ref{eq: 1-3 post-training}) by using teacher-provided signals as
post-training supervision. Let $\mathcal{D}_{\mathrm{dis}}$ denote the
post-training observations used for distillation, and let $s_T(\tilde{o})$
denote the teacher signal for observation $\tilde{o}$, such as a policy
distribution $\pi_{\text{teacher}}(\cdot
\mid\tilde{o})$ or teacher target $\hat y(\tilde o)$.
The student is then optimized by
\begin{equation}
\min_{\theta \in \mathcal{B}(\theta_0)}
\mathbb{E}_{\tilde{o}\sim \mathcal{D}_{\mathrm{dis}}}
\left[
\ell_{\mathrm{dis}}
\left(
\pi_\theta(\tilde{o}), s_T(\tilde{o})
\right)
\right]
+
\lambda\Omega(\theta),
\label{eq:distillation_post_training}
\end{equation}
where $\ell_{\mathrm{dis}}$ measures the mismatch between the student output and the teacher signal, and $\Omega(\theta)$ optionally regularizes the policy toward a stable reference. For example, $\ell_{\mathrm{dis}}$ can be a KL divergence for distribution matching or a regression/cross-entropy loss for pointwise teacher labels.

CRAFT \cite{craft2026} uses the policy itself as the teacher: it regularizes updates with a Kullback--Leibler (KL) term toward an exponential-moving-average copy, keeping the policy close to a stable reference during reward-driven refinement; here we focus on this distillation component, while its counterfactual reward design is discussed in Section~\ref{sec:reward}. PaIR-Drive \cite{pairdrive2026} uses expert demonstrations as the teacher signal, with a parallel imitation branch running alongside the RL branch. RAPID \cite{rapid2024} distills from an RL teacher and combines this with online policy adaptation. In addition, larger models can serve as teachers. VLM-AD~\cite{vlmad2026} distills VLM-generated reasoning and action annotations into an E2E policy through auxiliary heads. CoPhy \cite{cophy2026} distills the reasoning of a language-aligned model into the action policy. Found-RL \cite{foundrl2026} combines foundation-model distillation with RL to transfer high-level driving knowledge into a fast policy.

The \textbf{data} question is about how to construct the post-training observations \(\tilde{o}\) that receive teacher labels. For closed-loop driving, teacher targets are most useful when \(\tilde{o}\) corresponds to states visited by the student during rollout, rather than only states from a fixed expert dataset. This directly addresses the distribution shift and accumulating errors in closed-loop execution. The idea traces back to DAgger-style imitation: the student acts, a teacher labels the visited states, and the student is retrained on its own state distribution. CRAFT and PaIR-Drive follow this pattern, applying the teacher signal to the policy's own rollouts rather than to a fixed expert dataset.

In most systems, the teacher signal does not update the policy on its own. Instead, it constrains and stabilizes a reward-driven RL update, for example by adding a KL term that keeps the policy close to the teacher, or by letting the teacher generate and label rollouts. Compared with reward-based ranking, this denser supervision lets the model learn on states near the closed-loop distribution without a complex reward design, while the teacher serves as a stable reference that prevents the policy from drifting and losing the competence acquired in Supervised Fine-Tuning (SFT).

\section{Preference-based Alignment}
\label{sec:preference}

Preference-based alignment supervises the policy through relative comparisons between behaviors, which are easier to obtain in practice.
In autonomous driving, the key question in many scenarios is not whether a trajectory is feasible, but which one is more appropriate among multiple feasible choices. 

This family is closely related to RLHF, but it usually avoids training an explicit reward model. Instead, the policy is optimized directly from preference pairs. 
Formally, preference-based alignment instantiates Eq.~(\ref{eq: 1-3 post-training}) by using relative comparisons as post-training supervision. Let $\mathcal{D}_{\mathrm{pref}}=\{(\tilde{o}_i,y_i^w,y_i^l)\}_{i=1}^{M}$ denote the preference dataset, where $y_i^w$ is preferred over $y_i^l$ under observation $\tilde{o}_i$. A generic preference objective is
\begin{equation}
\min_{\theta \in \mathcal{B}(\theta_0)}
\mathbb{E}_{(\tilde{o},y^w,y^l)\sim \mathcal{D}_{\mathrm{pref}}}
\left[
\ell_{\mathrm{pref}}
\left(
\pi_\theta; \tilde{o}, y^w, y^l
\right)
\right]
+
\lambda\Omega(\theta),
\label{eq:preference_post_training}
\end{equation}
where $\ell_{\mathrm{pref}}$ penalizes the policy when the rejected behavior $y^l$ is assigned a higher preference score than the preferred behavior $y^w$, $\Omega(\theta)$ optionally regularizes the policy toward a reference policy, and $\lambda\ge 0$ controls the strength of this regularization. 

DPO~\cite{rafailov2023dpo} is the representative method: it updates the policy using the relative ordering between a preferred sample and a rejected sample. Following DPO, $\ell_{\mathrm{pref}}$ can be instantiated as
\[
-\log \sigma
\left(
\beta
\left[
\log
\frac{\pi_\theta(y^w\mid\tilde{o})}
{\pi_{\mathrm{ref}}(y^w\mid\tilde{o})}
-
\log
\frac{\pi_\theta(y^l\mid\tilde{o})}
{\pi_{\mathrm{ref}}(y^l\mid\tilde{o})}
\right]
\right),
\]
where $\sigma(\cdot)$ is the sigmoid function, $\pi_{\mathrm{ref}}$ is usually the imitation-learned or SFT policy and $\beta$ controls the strength of the preference update. Driving methods based on this idea mainly differ in how they construct preference pairs and how they adapt the loss to driving-specific objectives.

In this family, the preference pairs are both \textbf{the supervision and the data}, so their construction is the central design choice. DriveDPO~\cite{drivedpo2025} proposes Safety DPO, which combines imitation similarity with rule-based safety scores. In this way, the model learns trajectories that are close to human driving while avoiding behaviors that look human-like but are actually unsafe. VL-DPO~\cite{vldpo2026} uses a VLM to automatically generate preference pairs from driving rollouts, reducing the need for human annotation. CSN~\cite{csn2026} combines Plackett--Luce DPO with negative-log-likelihood regularization, allowing a VLA policy to learn human-like driving preferences from rankings over multiple candidate trajectories. CPO++~\cite{cpopp2026} injects counterfactual perturbations into reasoning and perception during preference optimization, so that spurious correlations are less likely to produce incorrect preferences. Other methods construct preference pairs from more specialized supervision sources. TakeAD~\cite{takead2025} builds preference pairs from expert takeover and disengagement events to further optimize an imitation policy toward better closed-loop behavior. Drive My Way~\cite{drivemyway2026} aligns a VLA policy with individual driving habits and natural-language intent for personalized driving.

The effectiveness of preference alignment depends critically on the informativeness of the selected trajectory pairs and the extent to which their comparisons reflect real-world human preferences. If the compared trajectories fail to expose meaningful behavioral trade-offs (with respect to objectives such as comfort or safety), the improvement of the refined policy may be moderate or limited. 

\section{Reinforcement Learning Post-training}
\label{sec:rl}

RL-based post-training for driving is inspired by the broader Large Language Model (LLM) post-training literature, including reward-model-based RLHF~\cite{ouyang2022instructgpt}, DPO~\cite{rafailov2023dpo}, and group-relative methods such as Group Relative Policy Optimization (GRPO)~\cite{shao2024deepseekmath,deepseek_r1_2025}. However, driving is not a direct transfer of these objectives. Driving quality is inherently multi-dimensional, long-horizon, and highly context-dependent, reflecting complex trade-offs among collision avoidance, drivable-area compliance, passenger comfort, and driving progress.

In RL-based post-training, supervision is provided by rewards assigned to sampled driving behaviors. These rewards may come from rule-based metrics, learned critics, world models, counterfactual comparisons, or reasoning-consistency checks.

Formally, let $\tau$ denote a sampled driving behavior, such as a trajectory, an action sequence, or a reasoning-action output, and let $R(\tau)$ denote the scalar reward or return assigned to it. A generic regularized RL objective can be written as
\begin{equation}
\min_{\theta \in \mathcal{B}(\theta_0)}
-
\mathbb{E}_{\tau\sim p_{\pi_\theta}(\tau)}
\left[
R(\tau)
\right]
+
\lambda\Omega(\theta),
\label{eq:rl_post_training}
\end{equation}
where $p_{\pi_\theta}(\tau)$ denotes the sampling or rollout distribution induced by the current policy $\pi_\theta$, $\Omega(\theta)$ optionally regularizes the updated policy toward a reference policy, and $\lambda\ge 0$ controls the strength of this regularization. When rewards are defined over multiple time steps, $R(\tau)$ may aggregate step-wise rewards, e.g., $R(\tau)=\sum_t \gamma^t r_t$, where $r_t$ is the reward at step $t$ and $\gamma$ is a discount factor.

Their usefulness depends on the rollouts used for training, so rollout generation, exploration, data diversity, and failure-scenario construction are central to RL post-training. The final design choice is how rewards over these rollouts are converted into policy updates, which depends on the policy output space, such as token sequences, continuous trajectories, diffusion outputs, or hybrid reasoning-action policies.

Operationally, RL post-training refines a competent base policy after pretraining or driving-specific SFT, sometimes followed by test-time scaling. We organize this section around the main design choices inside this stage: Section~\ref{sec:reward} reviews how rewards are constructed as supervision signals; Section~\ref{sec:exploration} discusses how informative training data and rollouts are collected through exploration, data diversity, and negative/recovery case construction; and Section~\ref{sec:algorithms} reviews how optimization algorithms convert these rewards and rollouts into policy updates across different policy spaces.

\begin{figure}[t]
    \centering
    \includegraphics[width=\linewidth]{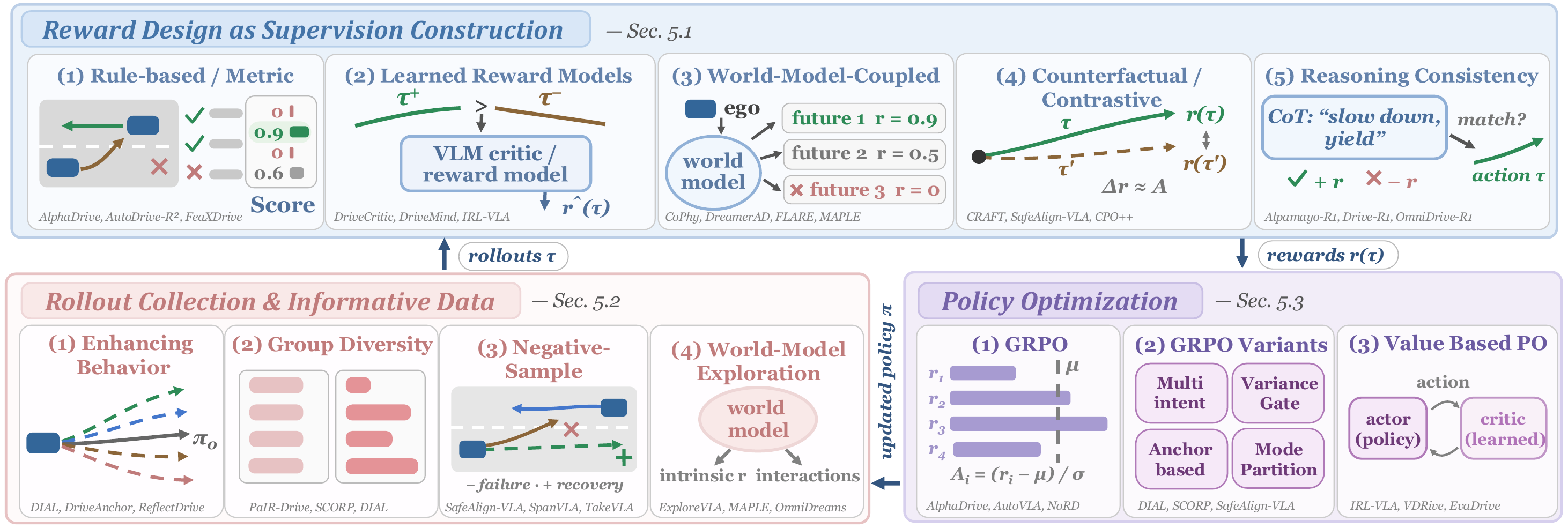}
    \vspace{-0cm}
    \caption{Overview of the RL design space for autonomous driving post-training.}
    \label{fig:fig3}
    \vspace{-0.5cm}
\end{figure}

\subsection{Reward Design as Supervision Construction}
\label{sec:reward}
In RL-based post-training, rewards serve as supervision by indicating which sampled behaviors should be reinforced. Unlike pointwise trajectory labels, driving rewards must capture multiple qualities at once, including safety, progress, comfort, rule compliance, and interaction. We therefore organize this subsection by how the reward signal is constructed.

\textbf{Rule-based and metric-derived rewards.}
Rule-based rewards convert observable rollout properties, such as collision, off-road driving, Time-to-Collision (TTC), progress, and smoothness, into scalar signals. They are cheap to compute, easy to reproduce, and widely used on standard benchmarks. One common form is to use benchmark scores directly, such as the NAVSIM Predictive Driver Model Score (PDMS)~\cite{navsim2024} and its Extended PDMS (EPDMS)~\cite{navsimv2_2025}, which combine collision avoidance, drivable-area compliance, TTC, comfort, and progress into a single reward. Another form is to hand-design weighted components, as in AutoDrive-R\textsuperscript{2}~\cite{autodriver2_2025} for spatial, dynamic, and temporal rewards, FeaXDrive~\cite{feaxdrive2026} for curvature feasibility, and PanoEnv~\cite{panoenv2026} for geometry-aware 3D spatial understanding. These rewards are interpretable, but fixed rules often miss interaction semantics, such as whether a lane change is natural or a yield is well timed. For this reason, they are often combined with learned rewards or guidance signals; for example, SARAD~\cite{sarad2026} pairs collision-prediction rewards with LLM guidance.

\textbf{Learned reward models.}
Learned reward models provide supervision for qualities that are difficult to write as fixed rules. They score candidate behaviors using learned critics, VLM evaluators, preference models, verifiers, or uncertainty signals. In VLA systems, VLM critics are especially natural: DriveMind~\cite{drivemind2025} uses a VLM for semantic anchoring, CoPhy~\cite{cophy2026} adds a language-aligned scorer, DriveCritic~\cite{drivecritic2025} trains a VLM evaluator from pairwise human preferences, and VDRive~\cite{vdrive2025} uses a learned critic in an actor–critic framework. This category also includes reward-model-based RLHF, where feedback is first used to train an explicit reward model before policy optimization. Sun et al.~\cite{sun2024rlhf} use physical and physiological feedback with Proximal Policy Optimization (PPO), while Zhuang et al.~\cite{neurocog2026} use EEG-derived signals. Learned rewards can also appear as verifiers or uncertainty signals: LaViPlan~\cite{laviplan2025} adapts Reinforcement Learning with Verifiable Rewards (RLVR) to graded trajectory assessment, ExploreVLA~\cite{explorevla2026} rewards world-model uncertainty, and IRL-VLA~\cite{jiang2025irl} learns a lightweight Reward World Model (RWM) from EPDMS-based metric labels before using it for PPO-based VLA fine-tuning.

\textbf{World-model-coupled rewards.}
World-model-coupled rewards use predicted future states to evaluate an action, since the quality of a lane change, yield, or braking decision depends on how the scene evolves. CoPhy~\cite{cophy2026} derives physical rewards from BEV rollouts and a cognitive scorer, DreamerAD~\cite{dreamerad2026} computes dense rewards in latent space, FLARE~\cite{flare2026} scores trajectories with future-aware VLM latents, and generative models such as OmniDreams~\cite{omnidreams2026} and MAPLE~\cite{maple2026} evaluate interactions between ego behavior, generated futures, and other agents. Across these methods, the world model is coupled with reward computation by extracting behavior-relevant signals from predicted futures with future-aware representations.

\textbf{Counterfactual and contrastive rewards.}
Counterfactual and contrastive rewards compare a trajectory with nearby alternatives, providing denser supervision when failures are rare or small action changes lead to different outcomes. CRAFT~\cite{craft2026} uses counterfactual advantages as a dense proxy for sparse closed-loop advantages, decomposing the signal into a proxy term and a grounded residual corrected by interaction-critical events. Related constructions also appear in SafeAlign-VLA’s safety pairs (Section~\ref{sec:exploration}) and CPO++’s counterfactual perturbations (Section~\ref{sec:preference}).

\textbf{Reasoning-quality and consistency rewards.}
For VLA policies, rewards can supervise not only the final trajectory but also the reasoning process behind it. Alpamayo-R1~\cite{alpamayo2025} penalizes explanations that mismatch the executed action, since a correct action with inconsistent reasoning may be only accidentally valid. Drive-R1~\cite{li2026drive} uses a GRPO reward that combines trajectory accuracy, meta-action correctness, repetition penalty, and format compliance. OmniDrive-R1~\cite{zhang2026omnidrive} further targets visual grounding with a CLIP-based Region-of-Interest (ROI) grounding reward. Other works assign rewards along the reasoning chain: AutoDrive-P\textsuperscript{3}~\cite{autodrivep3_2026} rewards perception, prediction, and planning separately, ELF-VLA~\cite{elfvla2026} uses failure-diagnostic feedback, and AutoDrive-R\textsuperscript{2}~\cite{autodriver2_2025}, ThinkDrive~\cite{thinkdrive2026}, and MindDriver~\cite{minddriver2026} couple Chain-of-Thought (CoT) reasoning with RL refinement. A separate line learns when to invoke CoT at all, as in AdaThinkDrive~\cite{adathinkdrive2025} and AutoVLA~\cite{autovla2025}. In contrast, NoRD~\cite{nord2026} shows that Dr.~GRPO can also be applied to VLA driving without explicit reasoning.

\subsection{Rollout Collection and Informative Data}
\label{sec:exploration}

\textbf{Informative Rollout Groups.}
RL post-training depends not only on the reward, but also on the rollouts used for training. In group-relative methods such as GRPO, learning comes from reward differences within a sampled group. If all candidates receive similar rewards, the advantages become small and the update provides little signal. This is especially common in driving. SFT policies tend to stay close to demonstrations, so reranking cannot recover trajectories that were never sampled~\cite{dial2026}. At the same time, routine scenes often lead to saturated rewards, making long-tail, interaction-heavy, and safety-critical cases more informative.

Useful rollout groups therefore need two properties. First, the policy must be able to sample different maneuvers, since RL cannot improve behaviors that never appear. DIAL~\cite{dial2026} conditions a flow head on intent labels with classifier-free guidance, DriveAnchor~\cite{driveanchor2026} uses a trajectory-anchor vocabulary, and ReflectDrive~\cite{reflectdrive2025} uses a discrete action codebook. These methods build diversity into the output space instead of relying only on random sampling. Second, the sampled candidates must differ in quality. PaIR-Drive~\cite{pairdrive2026} uses a tree-structured sampler over maneuvers, while SCORP~\cite{scorp2026} gates updates using within-group reward variance. Best-of-$N$ also provides a useful diagnostic: if performance saturates quickly as $N$ increases, the limitation is likely the sampling distribution rather than the reranker~\cite{dial2026}.

\textbf{Negative and Recovery Data Construction.}
Another way to obtain informative training data is to focus on negative and recovery cases. Such cases are rare in expert logs and hard to reach through naive exploration, but they provide strong learning signals by exposing what the policy cannot handle. SafeAlign-VLA~\cite{safealignvla2026} constructs positive and negative pairs from risky scenes for anchor-based GRPO. SpanVLA~\cite{spanvla2026} adds negative and recovery samples so the model learns both to avoid unsafe behavior and to correct mistakes. TakeVLA~\cite{takevla2026} mines human takeover data with Scenario Dreaming Reinforcement Fine-Tuning (RFT). Counterfactual-VLA~\cite{peng2026counterfactual} uses a rollout-filter-label pipeline to mine high-value rollout scenes and label counterfactual reasoning traces for action correction. ELF-VLA~\cite{elfvla2026} turns all-zero long-tail rewards into failure-diagnostic feedback over planning, reasoning, and execution.

\textbf{World-Model Rollouts.}
World models can evaluate imagined continuations under different actions, rather than only the future recorded in logs.  ExploreVLA~\cite{explorevla2026} uses world-model uncertainty to encourage exploration of uncertain states. MAPLE~\cite{maple2026} runs multi-agent latent rollouts to create interaction cases that are missing from the data. CoPhy~\cite{cophy2026} and OmniDreams~\cite{omnidreams2026} also use predicted futures to support both reward computation and rollout generation. These methods are most useful when the generated futures preserve the driving outcomes that matter for training, such as whether other agents react, whether the ego vehicle risks collision, and whether the interaction succeeds or fails.

\subsection{Policy Optimization Algorithms}
\label{sec:algorithms}
After rewards are assigned to rollouts, the optimization algorithm converts them into policy updates. This choice depends on the credit-assignment unit and policy output space, such as token sequences, continuous trajectories, flow-based generators, diffusion policies, or actor--critic architectures.

\textbf{Group-relative optimization for token-sequence policies.}
Many driving post-training methods formulate policy outputs as autoregressive token sequences, including action tokens, trajectory tokens, CoT rationales, and output formats. GRPO-style objectives fit this setting by sampling multiple completions for the same query, assigning each completion a task-specific reward, and computing group-normalized advantages without a separate value network. The same optimization template supports planning-oriented rewards over trajectory quality, as in AlphaDrive \cite{alphadrive2025}, and tokenized action or trajectory policies, as in AutoVLA \cite{autovla2025} and NoRD \cite{nord2026}. It also supports reasoning-oriented rewards over CoT consistency, process structure, or visual grounding, as in AutoDrive-P$^3$ \cite{autodrivep3_2026}, Drive-R1 \cite{li2026drive}, and OmniDrive-R1  \cite{zhang2026omnidrive}.

\textbf{Group-aware GRPO variants.}
Since group-relative advantages depend on which candidates are compared, several methods modify the group construction or the use of group statistics rather than only changing the reward. DIAL \cite{dial2026} uses multi-intent GRPO to construct intent-balanced candidate groups that span all predefined intent classes, SCORP \cite{scorp2026}
uses variance-gated GRPO, which gates sampled groups or changes
normalization according to within-group reward variance to reduce advantage
collapse and gradient instability, SafeAlign-VLA \cite{safealignvla2026} uses anchor-based GRPO over positive and negative risky-scene candidates, making the comparison set explicitly safety-focused,
DriveAgent-R1 \cite{zheng2025driveagent} further adapts this idea to hybrid reasoning policies through
Mode-Partitioned GRPO, forcing text-only and tool-augmented responses into the
same normalized group so that the update compares both within and across
reasoning modes.

\textbf{Sequence-level group-relative optimization beyond autoregressive decoding.}
Flow-GRPO \cite{liu2026flow} adapts group-relative policy optimization to flow-matching generators, where trajectory samples are produced through a denoising process rather than left-to-right token decoding. WAM-Diff \cite{xu2025wam} uses Group Sequence Policy Optimization (GSPO) framework which uses masked diffusion policy to refine the complete sequences of future trajectories.

\textbf{Value-based and reward-model-based policy gradients.}
Not all RL post-training methods use group-relative objectives. When a learned reward model or critic is available, PPO or actor-critic updates become natural.
IRL-VLA \cite{jiang2025irl} learns an RWM and then uses PPO to fine-tune the VLA policy under learned reward feedback. VDRive \cite{vdrive2025} trains its diffusion policy head in an actor--critic framework, using a learned critic to guide policy optimization.

\section{Test-time Refinement}
\label{sec:testtime}

Test-time refinement refers to improving model outputs during inference without updating model parameters. It does not retrain the policy; instead, 
the policy $\pi_{\theta_0}$ is kept fixed and refinement is performed over its generated outputs. Following Eq.~(\ref{eq:test_time_selection}), given an observation $o$, a candidate set $\mathcal{Y}_{\mathrm{cand}}(o)=\{y_i\}_{i=1}^{K}$ can be sampled from the fixed policy with $y_i\sim\pi_{\theta_0}(\cdot\mid o)$, and a verifier or reranker $v(o,y)$ selects the final output $y^*$.

The most common approach is to first generate multiple candidate trajectories and then use a scoring model to select the best one. The human-aligned evaluator in DriveCritic \cite{drivecritic2025} can be used as a reranker during inference. However, Best-of-$N$ is useful only when the sampled candidates already contain a good trajectory; if the base policy produces only low-quality candidates, the reranker can merely choose among poor options, and performance quickly saturates~\cite{dial2026}. Therefore, test-time refinement is better understood as a complement to other post-training methods: it can help a strong policy select better outputs, but it cannot truly learn behaviors that the policy has never generated.

Another type of method is self-reflection, where the model first generates an initial decision, then checks whether it contains unsafe or unreasonable parts and attempts to revise them. ReflectDrive \cite{reflectdrive2025} represents driving behavior with a discrete action codebook and improves the output by searching for and replacing unsafe action tokens. AutoDrive-R\textsuperscript{2} \cite{autodriver2_2025} incorporates self-reflection into its reasoning policy. C-CoT \cite{ccot2026} further introduces counterfactual CoT reasoning at inference time. This allows the VLM to consider the possible outcomes of different choices before making a final decision in high-risk scenarios. Seong et al.~\cite{grbo2025} combine RL post-training of an interactive behavior model with test-time scaling. Their results show that test-time refinement can complement parameter-updating methods rather than replace them.

\textbf{On-vehicle efficiency constraints.} Compared with test-time scaling in language models, test-time refinement in autonomous driving remains less explored. One important reason is that on-vehicle latency constraints are strict, so inference cannot afford arbitrary extra computation. Therefore, how to perform more effective candidate generation, search, and reranking under a limited compute budget is a key problem for this direction. The block-diffusion decoding in Fast-dDrive \cite{fastddrive2026} is an efficiency-oriented design along this line.

\section{Evaluation Protocols}
\label{sec:evaluation}

\input{benchmark-table-2}

\subsection{Open-loop Evaluation}
\label{sec:eval:openloop}

Evaluation protocols for autonomous driving post-training range from open-loop testing on recorded data to fully closed-loop simulation. In open-loop evaluation, the model is tested on fixed scenarios, and its actions do not affect future states. Common benchmarks include nuScenes~\cite{nuscenes2020} and the Waymo Open Dataset for End-to-End Driving (WOD-E2E)~\cite{wode2e2025}. Typical metrics include Average and Final Displacement Error (ADE/FDE), collision rate, and the Rater Feedback Score (RFS) in WOD-E2E. Open-loop evaluation is inexpensive, reproducible, and useful for measuring how well models fit logged trajectories or human preference annotations.

Table~\ref{tab:openloop} summarizes open-loop results reported by post-trained policies on nuScenes and WOD-E2E. However, open-loop evaluation cannot capture how the policy's actions change future states. It therefore cannot measure accumulating errors, multi-agent interaction, or recovery ability. This is especially limiting for post-training, whose goal is to improve closed-loop behavior. Thus, open-loop scores should be viewed as auxiliary indicators rather than final evidence of driving quality. Zhao et al.~\cite{bridgesim2026} further show that the open-loop-to-closed-loop gap comes from observational domain shift and objective mismatch, and that standard open-loop protocols can hide closed-loop failures.

\subsection{Pseudo-Closed-loop and Closed-loop Evaluation}
\label{sec:eval:closedloop}

Closed-loop evaluation is most aligned with post-training because the policy's actions affect future states. In practice, common benchmarks vary in how closed-loop they are. NAVSIM v1~\cite{navsim2024} is a one-shot, non-reactive simulation: the policy plans once from recorded observations, and the plan is unrolled for a few seconds. Since the policy receives no feedback and the scene does not react, it is best viewed as open-loop evaluation with simulation-based metrics. Its PDMS score combines collision, drivable area, ego progress, comfort, and TTC. NAVSIM v2~\cite{navsimv2_2025} moves closer to closed-loop evaluation with a two-stage pseudo-simulation protocol, where synthetic follow-up observations depend on the first-stage behavior. Its EPDMS metric extends PDMS with driving-direction compliance, traffic-light compliance, lane keeping, and extra comfort terms, and the harder navhard split is evaluated under this protocol.

Bench2Drive~\cite{bench2drive2024} is fully closed-loop. It is built on CARLA with reactive multi-agent traffic and reports Driving Score (DS) and Success Rate (SR), making it better suited for evaluating long-horizon errors and interaction behavior. Recent work also moves closed-loop testing beyond game engines. AlpaSim~\cite{alpasim2025} uses neural rendering for photorealistic sensor simulation, enabling closed-loop testing on realistic sensor streams. In addition, nuReasoning~\cite{nureasoning2026} focuses on long-tail reasoning scenarios. Its nuReasoning Planning Score (NPS) follows a NAVSIM-style safety-gated score and is reported with 5-second ADE. Table~\ref{tab:benchmark} collects the results reported on NAVSIM and Bench2Drive, grouped by whether the policy uses a foundation-model backbone.

\input{benchmark-table-1}
\vspace{-0.5cm}

\subsection{Main Challenges in Current Evaluation}
\label{sec:eval:challenges}

Closed-loop evaluation is better aligned with the goal of post-training than open-loop evaluation, but current protocols still have several limitations.

\textbf{Benchmark saturation.} NAVSIM has become the default evaluation platform for many recent works. As Table~\ref{tab:benchmark} shows, PDMS and EPDMS scores now cluster within a narrow range, making a single headline score less useful for separating real methodological progress from random variation. Although PDMS already includes sub-scores such as collision, drivable-area compliance, TTC, comfort, and progress, many methods perform similarly on these components in routine scenes. As a result, small score differences are often hard to interpret. More discriminative evaluation should therefore focus on harder distributions, such as long-tail, safety-critical, and interaction-heavy splits like navhard, where component-level differences are more likely to remain visible.

\textbf{Scarce and non-comparable real-world evaluation.} Real-vehicle testing requires vehicle platforms, safety drivers, and test permits, which are difficult for most academic groups to access. Still, some recent works have started to move beyond benchmark evaluation. For example, Alpamayo-R1~\cite{alpamayo2025} reports deployment latency, HDP~\cite{hdp2026} reports real-road testing, DriveAnchor~\cite{driveanchor2026} and RAD-2~\cite{rad2_2026} report real-vehicle validation, and VLM-AutoDrive~\cite{vlmautodrive2026} evaluates safety-critical event detection on real dashcam videos. However, these results are still difficult to compare across papers because the vehicle platforms, sensors, scenarios, and metrics differ.

\textbf{Insufficient reporting of inference cost.} Post-training gains have limited practical value if the model cannot meet real-time on-vehicle constraints, yet many papers still omit inference cost. Unlike real-vehicle testing, this can be measured without access to a vehicle. Latency, parameter count, and throughput can be profiled on automotive-grade hardware, as shown by Alpamayo-R1 with 99 ms end-to-end latency, DriveAnchor with 2.06 ms inference time, and BPF~\cite{bpf2026} with up to 27\% higher decoding throughput on Jetson AGX Orin through pruning during RL. These results suggest that on-vehicle efficiency should also be reported alongside closed-loop performance.

\section{Open Problems and Future Directions}
\label{sec:future}

Post-training for autonomous driving remains an early-stage area, and its most important open problems cut across the four families instead of belonging to any single one.

\textbf{Reward and exploration as a coupled design problem.} 
Most RL-based post-training methods combine several reward sources, such as rule-based scores, learned critics, world-model rollouts, and reasoning-consistency terms. However, the way these rewards are combined is often under-studied: weights are usually hand-tuned, gating rules are not always reported, and a composite score may improve by exploiting an easy component rather than improving overall driving quality. It may also hide trade-offs among safety, progress, comfort, and rule compliance, so score gains should be analyzed at the component level. 

Reward design is also tied to exploration. A reward can guide learning only when sampled rollouts show meaningful quality differences. If all candidates in a group receive similar scores, group-relative advantages become small or vanish, and the update provides little signal. Thus, the goal is not diversity for its own sake, but reward-relevant diversity: sampled behaviors should differ in ways that the reward can reliably distinguish. Co-designing reward and exploration so that rewards reflect driving quality and rollout groups remain informative is a central open problem for RL-based post-training.

\textbf{Are world-model signals behavior-relevant?}
World models are increasingly used as both reward providers (Section~\ref{sec:reward}) and exploration engines (Section~\ref{sec:exploration}). Their value for post-training is not only visual realism, but also counterfactual feedback: they can estimate what would happen under actions that were not taken in the logged data. This gives post-training access to outcome-based signals that static supervision cannot provide.

However, many world models operate at the pixel or sensor level, while post-training needs behavior-level signals such as safety, progress, interaction, and rule compliance. For policy learning, the key requirement is therefore not pixel-level fidelity alone, but behavior-level fidelity: the model should preserve the factors that affect driving rewards, such as collisions, agent reactions, route progress, and traffic-rule violations. World-model rewards may be most useful when combined with rule-based or learned-critic signals, where the world model provides counterfactual future states and explicit reward terms anchor optimization to driving objectives.

\textbf{From one-off refinement to data-feedback loops.}
Post-training depends on the data and feedback signals available after the initial model is trained. In autonomous driving, the most useful signals are often not ordinary logged trajectories, but cases that reveal model weaknesses: takeovers, disengagements, near-misses, rule violations, recovery behaviors, and difficult interactions. Recent works such as TakeAD~\cite{takead2025}, TakeVLA~\cite{takevla2026}, and SpanVLA~\cite{spanvla2026} begin to use such field signals or recovery-oriented data to improve post-training.

However, these signals are still mostly used offline, as a fixed dataset for a single post-training stage. What is missing is a more systematic feedback loop: deployed or evaluated policies should expose failures, these failures should be converted into usable supervision, preferences, rewards, or recovery targets, and the refined policy should then be re-evaluated on the same types of hard cases. Building such data-feedback loops may be essential for making post-training a continual improvement process rather than a one-off refinement step.

\textbf{Beyond GRPO-based optimization.}
Most current RL-based post-training methods for autonomous driving use GRPO-based optimization, largely following recent practice in large language model alignment. While this is a useful starting point, driving has more structured learning signals: a trajectory is shaped by time, safety, interaction, and physical feasibility. Future methods may therefore need more driving-aware updates. Instead of assigning one scalar reward to the whole rollout, they could use more localized feedback around key moments, such as lane changes, yielding, braking, or near-collision events. They could also separate safety, progress, and comfort constraints, since improvements in one dimension may hide degradation in another. 

More broadly, post-training does not have to rely only on RL. Distillation can provide dense teacher targets from planners, stronger VLMs, or corrected rollouts, especially for failure and recovery cases. Preference-based alignment can compare alternative trajectories when the desired behavior is hard to express as a pointwise label. The key question is not which algorithm family is best, but which learning signal best matches each driving problem.

\section{Conclusion}
\label{sec:conclusion}

We presented a unified view of post-training for autonomous driving: a stage applied after an initial policy has been obtained, whose objective is closed-loop behavior rather than offline fit. Organizing the literature by the form of supervision yields four families: distillation, preference-based alignment, RL, and test-time refinement. Within the most developed family, RL, the central design questions are reward construction and exploration rather than the optimizer, while evaluation is becoming the bottleneck as closed-loop benchmarks saturate. We hope this survey offers a coherent reference and that its taxonomy and open problems help structure future work.

\bibliographystyle{splncs04}
\bibliography{references}

\end{document}

%% file: benchmark-table-2.tex
\begin{table}[t]
\centering
\caption{Open-loop results on nuScenes and WOD-E2E. L2/ADE: displacement error; Col.: collision rate; RFS: human rater feedback score.}
\vspace{-0.2cm}
\label{tab:openloop}
\scriptsize
\setlength{\tabcolsep}{5pt}
\renewcommand{\arraystretch}{1.1}
\resizebox{\textwidth}{!}{%
\begin{tabular}{@{}l cc ccc C{4.0cm}@{}}
\toprule
\multicolumn{1}{@{}c}{\multirow{2}{*}{\textbf{Method}}}
& \multicolumn{2}{c}{\textbf{nuScenes}}
& \multicolumn{3}{c}{\textbf{Waymo E2E (WOD-E2E)}}
& \multicolumn{1}{c@{}}{\multirow{2}{*}{\textbf{Backbone}}} \\
\cmidrule(lr){2-3}\cmidrule(lr){4-6}
& {L2\dn} & {Col.\dn} & {RFS\up} & {ADE 5s\dn} & {ADE 3s\dn} & \\
\midrule
AutoDrive-R2 \cite{autodriver2_2025}  & \best{0.19}  & 0.07 & \nr   & \nr   & \nr   & Qwen2.5-VL-7B \\
Reasoning-VLA \cite{reasoningvla2025} & 0.23 & 0.08 & \nr   & \nr   & \nr   & Qwen2.5-VL-7B \\
VDRive \cite{vdrive2025}              & 0.29 & 0.18 & \nr   & \nr   & \nr   & InternVL3-8B \\
VLA-World \cite{vlaworld2026}         & 0.30 & 0.10 & \nr   & \nr   & \nr   & Qwen2-VL-2B \\
Drive-R1 \cite{li2026drive}           & 0.31 & 0.09 & \nr   & \nr   & \nr   & InternVL2-4B \\
LLaViDA \cite{llavida2025}            & 0.31 & 0.10 & \nr   & \nr   & \nr   & LLaVA-NeXT-8B \\
Fast-dDrive \cite{fastddrive2026}     & 0.32 & \nr  & 7.823 & 2.907 & 1.254 & Qwen2.5-VL-3B \\
AutoDrive-P3 \cite{autodrivep3_2026}  & 0.33 & \best{0.06} & \nr   & \nr   & \nr   & Qwen2.5-VL-3B \\
MindDriver \cite{minddriver2026}      & 0.33 & 0.12 & \nr   & \nr   & \nr   & Qwen2.5-VL-3B \\
LaST-VLA \cite{lastvla2026}           & 0.38 & 0.18 & \nr   & \nr   & \nr   & InternVL3-8B \\
AutoVLA \cite{autovla2025}            & 0.40 & 0.20 & 7.557 & 2.958 & 1.351 & Qwen2.5-VL-3B \\
ExploreVLA \cite{explorevla2026}      & 0.44 & 0.10 & \nr   & \nr   & \nr   & Show-o-1.3B \\
Poutine \cite{poutine2025}            & \nr  & \nr  & \best{7.990} & \best{2.740} & \best{1.210} & Qwen2.5-VL-3B \\
NoRD \cite{nord2026}                  & \nr  & \nr  & 7.709 & 2.893 & 1.250 & Qwen2.5-VL-3B \\
\bottomrule
\end{tabular}%
}
\end{table}

%% file: benchmark-table-1.tex
\begin{table}[t]
\centering
\caption{Reported results for post-trained driving policies. NAVSIM: PDMS/EPDMS; Bench2Drive: Driving Score (DS) and Success Rate (SR).}
\label{tab:benchmark}
\scriptsize
\setlength{\tabcolsep}{4pt}
\renewcommand{\arraystretch}{1.1}
\resizebox{\textwidth}{!}{%
\begin{tabular}{@{}l ccc cc C{4.3cm}@{}}
\toprule
\multicolumn{1}{@{}c}{\multirow{2}{*}{\textbf{Method}}}
& \multicolumn{3}{c}{\textbf{NAVSIM}}
& \multicolumn{2}{c}{\textbf{Bench2Drive}}
& \multicolumn{1}{c@{}}{\multirow{2}{*}{\textbf{Backbone}}} \\
\cmidrule(lr){2-4}\cmidrule(lr){5-6}
& {v1\up} & {v2\up} & {navhard\up} & {DS\up} & {SR\up} & \\
\midrule
\rowcolor{groupbg}\multicolumn{7}{@{}l}{\textbf{\textit{With Foundation Model}}}\\
Reasoning-VLA \cite{reasoningvla2025}  & 91.7 & \nr  & \nr  & \nr   & \nr   & Qwen2.5-VL-7B \\
FLARE \cite{flare2026}                 & 91.4 & 86.3 & \nr  & \nr   & \nr   & Qwen3-VL-4B \\
LaST-VLA \cite{lastvla2026}            & 91.3 & 87.1 & \nr  & \nr   & \nr   & InternVL3-8B \\
ReflectDrive \cite{reflectdrive2025}   & 91.1 & \nr  & \nr  & \nr   & \nr   & LLaDA-V-8B \\
ELF-VLA \cite{elfvla2026}              & 91.0 & 87.1 & \nr  & \nr   & \nr   & InternVL3-8B \\
AutoDrive-P3 \cite{autodrivep3_2026}   & 90.6 & \best{89.9} & \nr & \nr & \nr & Qwen2.5-VL-3B \\
ExploreVLA \cite{explorevla2026}       & 90.4 & 88.8 & \nr  & \nr   & \nr   & Show-o-1.3B \\
EponaV2 \cite{eponav2_2026}            & 90.4 & 88.9 & \best{52.0} & \nr & \nr & Qwen3-VL-4B \\
SpanVLA \cite{spanvla2026}             & 90.3 & 86.4 & 40.1 & \nr   & \nr   & Qwen2.5-VL-3B \\
AdaThinkDrive \cite{adathinkdrive2025} & 90.3 & \nr  & \nr  & \nr   & \nr   & InternVL3-8B \\
FeaXDrive \cite{feaxdrive2026}         & 90.0 & \nr  & \nr  & \nr   & \nr   & InternVL3-2B \\
AutoVLA \cite{autovla2025}             & 89.1 & \nr  & \nr  & 78.84 & 57.73 & Qwen2.5-VL-3B \\
AutoDrive-R2 \cite{autodriver2_2025}   & 89.1 & \nr  & \nr  & \nr   & \nr   & Qwen2.5-VL-7B \\
SafeAlign-VLA \cite{safealignvla2026}  & 89.1 & \nr  & \nr  & \nr   & \nr   & Qwen2.5-VL-7B \\
NoRD \cite{nord2026}                   & 85.6 & \nr  & \nr  & \nr   & \nr   & Qwen2.5-VL-3B \\
TakeVLA \cite{takevla2026}             & \nr  & \nr  & \nr  & \best{89.72} & \best{73.73} & Qwen2-0.5B \\
MAPLE \cite{maple2026}                 & \nr  & \nr  & \nr  & 88.30 & 70.50 & Qwen2.5-VL-1.5B \\
Drive My Way \cite{drivemyway2026}     & \nr  & \nr  & \nr  & 82.72 & 71.56 & InternVL2-1B \\
CoReVLA \cite{corevla2025} & \nr  & \nr  & \nr  & 72.18 & 50.00 & Qwen2.5-VL-7B \\
VDRive \cite{vdrive2025}               & \nr  & \nr  & \nr  & 66.25 & 50.51 & InternVL3-8B \\
MindDriver \cite{minddriver2026}       & \nr  & \nr  & \nr  & 65.48 & 39.55 & Qwen2.5-VL-3B \\
\addlinespace[1pt]
\midrule
\rowcolor{groupbg}\multicolumn{7}{@{}l}{\textbf{\textit{Without Foundation Model}}}\\
EvaDrive \cite{evadrive2025}    & \best{94.9} & 86.3 & \nr & 64.96 & 40.45 & -- \\
CoPhy \cite{cophy2026}          & 91.4 & 86.1 & \nr  & \nr   & \nr   & -- \\
PaIR-Drive \cite{pairdrive2026} & 91.2 & 87.9 & \nr  & \nr   & \nr   & -- \\
DriveDPO \cite{drivedpo2025}    & 90.0 & \nr  & \nr  & 62.02 & 30.62 & -- \\
TSA-MPR \cite{arplanning2025}   & 89.8 & \nr  & \nr  & \nr   & \nr   & -- \\
DreamerAD \cite{dreamerad2026}  & 88.7 & 87.7 & \nr  & \nr   & \nr   & -- \\
CRAFT \cite{craft2026}          & \nr  & \nr  & \nr  & 75.79 & 56.36 & -- \\
TakeAD \cite{takead2025}        & \nr  & \nr  & \nr  & 71.39 & 40.83 & -- \\
\bottomrule
\end{tabular}%
}
\end{table}